\documentclass[letterpaper, 10 pt, conference]{ieeeconf}  

\IEEEoverridecommandlockouts                              

\usepackage{graphics} 
\usepackage{epsfig} 
\usepackage{mathptmx} 
\usepackage{times} 
\usepackage{amsmath} 
\usepackage{amssymb}  

\usepackage[utf8]{inputenc}
\usepackage{pgfplots} 
\usepackage{pgfgantt}
\usepackage{pdflscape}
\pgfplotsset{compat=newest} 
\pgfplotsset{plot coordinates/math parser=false}
\usepackage{tikz}

\usepackage{cite}
\usepackage{amsmath,amssymb,amsfonts}
\usepackage{algorithm, algorithmic}
\usepackage{graphicx}
\usepackage{textcomp}
\usepackage{xcolor}
\usepackage{svg}
\usepackage{array}
\usepackage{booktabs}
\usepackage{optidef}
\usepackage{afterpage}
\usepackage{bm}
\usepackage{gensymb}
\usepackage{colortbl}
\usepackage{color}
\usepackage{tabularx}
\usepackage{hhline}
\usepackage{makecell}
\usepackage{hyperref}
\usepackage{wrapfig}

\let\labelindent\relax
\usepackage{enumitem}

\usepackage{pgfplots}
\pgfplotsset{compat=newest}
\usepackage{tikzscale}

\usepackage{pgf-pie} 
\usepackage{nicefrac}

\definecolor{green}{rgb}{0.3529, 0.7451, 0.4824}
\definecolor{greenish}{rgb}{0.6941, 0.8353, 0.502}
\definecolor{yellow}{rgb}{1, 0.9216, 0.5176}
\definecolor{orange}{rgb}{0.9882, 0.749, 0.4824}
\definecolor{lred}{rgb}{0.9725, 0.4118, 0.4196}

\definecolor{color1}{rgb}{0.39216,0.56078,1.00000}%
\definecolor{color2}{rgb}{0.4705,0.36863,0.9411764705882353}%
\definecolor{color3}{rgb}{0.86275,0.14902,0.49804}%
\definecolor{color4}{rgb}{0.99608,0.38039,0.00000}%
\definecolor{color5}{rgb}{1,0.6901960784313725,0.00000}%

\colorlet{green}{color2!80!white}
\colorlet{greenish}{color2!60!white}
\colorlet{yellow}{color2!40!white}
\colorlet{orange}{color2!20!white}
\colorlet{lred}{color2!0!white}

\colorlet{lblue}{color1!50!white}

\colorlet{green}{color2}
\colorlet{greenish}{color2!60!lblue}
\colorlet{yellow}{color2!20!lblue}
\colorlet{orange}{lblue!50!white}
\colorlet{lred}{lblue!0!white}

\usepackage{makecell}

\def\*#1{\textbf{#1}}

\title{\LARGE \bf

phloSAR: a Portable, High-Flow Pressure Supply and Regulator Enabling Untethered Operation of Large Pneumatic Soft Robots

}

\author{Maxwell Ahlquist*, Rianna Jitosho*, Jiawen Bao, and Allison M. Okamura
\thanks{*These authors contributed equally to this work.}%
\thanks{This work was supported in part by National Science Foundation (NSF) grant 2024247, an NSF Graduate Research Fellowship, the U.S. Department of Energy, National Nuclear Security Administration, Office of Defense Nuclear Nonproliferation Research and Development (DNN R\&D) under subcontract from Lawrence Berkeley National Laboratory; and the United States Federal Bureau of Investigation contract 15F06721C0002306.}
\thanks{Department of Mechanical Engineering, Stanford University. Email: \{ahlquist, rjitosho, jiawenb, aokamura\}@stanford.edu}%
}

\begin{document}

\maketitle
\thispagestyle{empty}
\pagestyle{empty}

\begin{abstract}
Pneumatic actuation benefits soft robotics by facilitating compliance, enabling large volume change, and concentrating actuator weight away from the end-effector. However, portability is compromised when pneumatic actuators are tethered to cumbersome air and power supplies. While there are existing options for portable pneumatic systems, they are limited in dynamic capabilities, constraining their applicability to low pressure and/or small-volume soft robots. In this work, we propose a portable, high-flow pressure supply and regulator (phloSAR) for use in untethered, weight-constrained, dynamic soft robot applications. PhloSAR leverages high-flow proportional valves, an integrated pressure reservoir, and Venturi vacuum generation to achieve portability and dynamic performance. We present a set of models that describe the system dynamics, experimentally validate them on physical hardware, and discuss the influence of design parameters on system operation. Lastly, we integrate a proof-of-concept prototype with a soft robot arm mounted on an aerial vehicle to demonstrate the system's applicability to mobile robotics. Our system enables new opportunities in mobile soft robotics by making untethered pneumatic supply and regulation available to a wider range of soft robots.

\end{abstract}

\section{Introduction} 
\label{sec:intro}
Soft robots exhibit passive adaptation to their environment, enabling inherent safety and robustness. Pneumatic actuation is a common choice for soft robots as it facilitates compliance and large volume change while relocating actuator mass away from the end-effector. However, pneumatic soft robots (PSRs) require a method for pressure generation and regulation, and many of these robots rely on pneumatic tubing that tethers them to heavy air compressors~\cite{jumet2022review}.
Meanwhile, untethered PSRs leverage methods such as micro pumps~\cite{Shtarbanov2021_FlowIO, Marcin2019_Board, Holland2014_Toolkit, Oguntosin2018_LowCost, Shrivastava2019_ProgrammableAir}, syringe displacement systems~\cite{Wu2020_Syringe}, CO$_2$ gas cartridges~\cite{Marchese2013, electronics_free}, and chemical reactions~\cite{chemical}. However, these untethered solutions are intended for PSRs with smaller volumes or low-flow requirements. While untethered pneumatic solutions exist for larger PSRs~\cite{Walsh, tolley2014}, this comes at the cost of added size and weight, limiting their suitability in PSR applications with stringent payload limits (e.g., soft aerial manipulation). 

In this work, we address the limitations of existing solutions regarding performance and portability with a portable, high-flow, pressure supply and regulator (phloSAR).
It leverages high-flow pneumatic components for high-speed pressure regulation of large volumes, and it integrates a refillable high-pressure air reservoir that enables untethered operation of PSRs.
This work provides the following contributions:
\begin{enumerate}
    \item The design details of phloSAR, which uses high-flow proportional valves, an integrated pressure reservoir, and Venturi vacuum generation, enabling untethered operation of large-volume pneumatic soft robots.
    \item Models, verification, and guidelines that aid in the design of future phloSARs for custom applications.
    \item Implementation of a physical phloSAR prototype and a demonstration on a mobile soft robot.
\end{enumerate}
\begin{figure}[t]
    \centering
    \includegraphics[width=\columnwidth, trim={0 10mm 0 7mm}, clip]{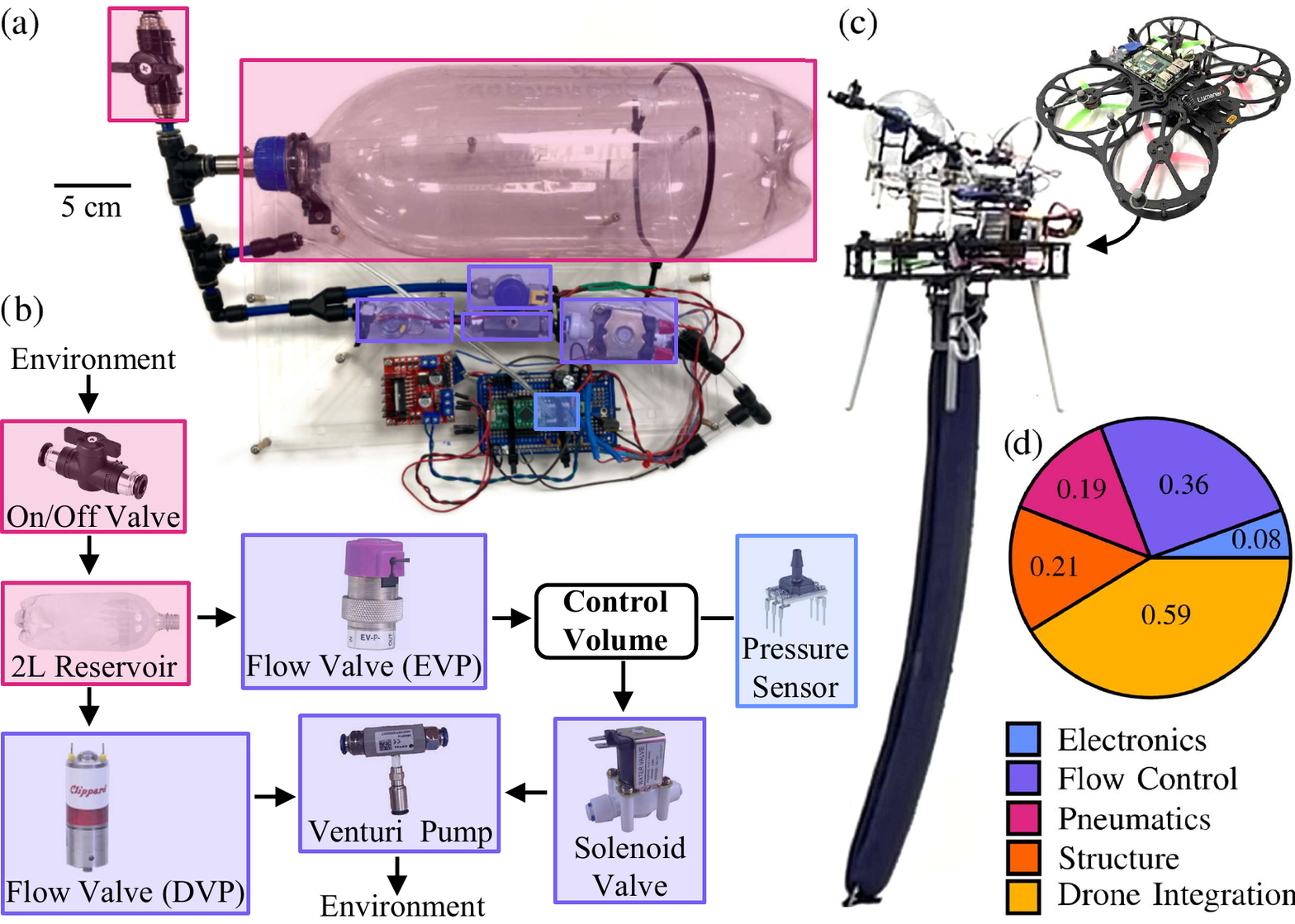}
    \vspace{-5mm}
    \caption{We present phloSAR, a portable high-flow pressure supply and regulator to enable untethered operation of large pneumatic soft robots. (a) PhloSAR prototype. (b) PhloSAR components and direction of airflow. (c) The prototype is integrated with an aerial vehicle and actuates a soft ``vine" robot to demonstrate the phloSAR portability and that it can be used to realize mobile soft robots. (d) Mass distribution (chart values in kg). ``Flow Control" includes the valves and Venturi pump, ``Pneumatics" includes the reservoir and pneumatic connections, ``Structure" includes the chassis and mounting hardware, and ``Drone Integration" includes the vine robot and additional mounting hardware. The phloSAR mass is 0.84~kg. The total mass (phloSAR + soft robot) is 1.4~kg, which is within the drone payload capacity (approximately 1.5~kg).}
    \vspace{-6mm}
    \label{fig:splash}
\end{figure}
\begin{figure}[t]
\begin{center}
    \vspace{.7mm}
    \input{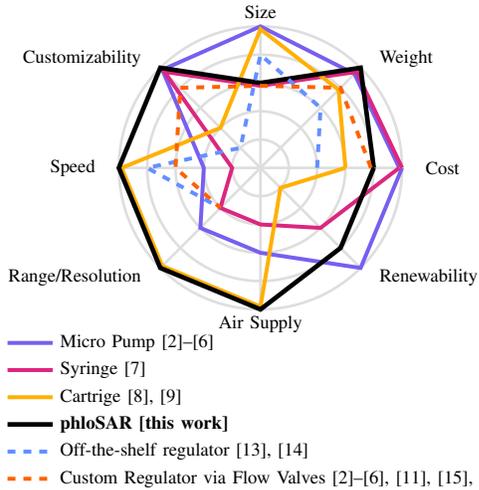}
    \vspace{-4mm}
    \caption{Qualitative comparison of existing pressure source and/or regulation solutions. Metrics are written around the circumference, and higher radii correspond with higher quality. Off-the-shelf regulators and custom regulators as is are not ready for untethered operation (they must be paired with a separate pressure source) and thus do not have quality values for ``Air Supply" and ``Renewability".}
    \label{fig:compass}
    \vspace{-8mm}
\end{center}
\end{figure}


This work aims to be a design guide to aid in implementing a phloSAR given a specific mobile PSR. As a motivating example, the phloSAR implemented in this work was customized for an inflated-beam ``vine" robot \cite{HawkesScienceRobotics2017}. These continuum robots grow in length via tip eversion and often feature pneumatic actuation for bending \cite{BlumenscheinFrontiers2020}. Traditional vine robots operate from a base fixed in the environment and require pneumatic tethering to heavy air compressors~\cite{CoadRAM2020, roboa}. In this work, we demonstrate that a vine robot with a phloSAR is portable enough to operate onboard an aerial vehicle, opening applications in aerial manipulation not previously feasible for this class of soft robot (Fig.~\ref{fig:splash}).

\section{Design Considerations and Prior Work}
\label{sec:prior_work}
When considering pneumatic solutions for mobile soft robots, three key design factors are its form, dynamic performance, and amenability to untethered operation. For form, we consider cost, weight, size, and customizability. For dynamic performance, we consider speed of pressure change and the output pressure range and resolution. For amenability to untethered operation, we consider whether the solution incorporates a portable pressure source, and whether the choice of air supply impacts system redeployment or operation duration (which we term ``renewability"). An overview can be seen in Fig~\ref{fig:compass}.

These design choices are inter-dependent, and the specific balance of requirements will be informed by the PSR for which the pneumatic system is intended.

\subsection{Dynamic Performance Considerations}
\label{sec:dynamic-considerations}
A pneumatic soft robot or actuator can be considered as a sealed volume requiring pressure regulation; this paper uses the general term ``control volume" (CV). To understand the dynamics of pressurizing an arbitrary control volume, we consider the ideal gas law in its derivative form:\vspace{-1.5mm}
\begin{equation}
    P_{cv}V_{cv} = n_{cv}R_uT \quad \rightarrow \quad \dot P_{cv} = \frac{\rho R_u T}{M V_{cv}}Q,\\[-1mm]
    \label{eq:pressure-rate-to-flow}
\end{equation}
where $\dot P_{cv}$ is the rate of pressure increase of the CV, $Q$ is the standard volumetric flow rate into the CV, $R_u$ is the universal gas constant, $n_{cv}$ is the number of gas moles in the CV, and $\rho$, $T$, $M$, are the density, temperature, and molar mass of air at standard conditions, respectively. From this relationship, we see that high-speed CV pressurization requires a high-flow (high $Q$) regulator. To understand the factors that determine flow rate, we consider Ohm's Law of fluid flow:\vspace{-1.5mm}
\begin{equation}
    \Delta P = QR_v,\\[-1mm]
    \label{eq:ohm_analogy}
\end{equation}
where $\Delta P$ is pressure difference across a valve and $R_\text{v}$ is the effective flow resistance across a valve (units can be derived from dimensional analysis). In the context of pressure supply and regulation, a pneumatic system with strong dynamic performance features a large pressure difference between the pressure supply and CV ($\Delta P$) and operates with a small flow resistance ($R_\text{v}$). 

\subsection{Untethered Pressure Sources}
Untethered operation of a PSR is typically dependent on having a portable pressure source. 
The two main options for this are (1) devices that generate airflow, and (2) pressure reservoirs. 



A common device for generating airflow for untethered PSRs is a micropump~\cite{Shtarbanov2021_FlowIO, Marcin2019_Board, Holland2014_Toolkit, Oguntosin2018_LowCost, Shrivastava2019_ProgrammableAir, tolley2014}. These devices exist in small, lightweight form factors and can continuously supply pressure as long as they have battery power. However, their flow rate drops as pressure accumulates at the outlet, and they have limited maximum flow rates up to approximately 10 standard liters per minute (SLPM). While these limitations do not preclude small-volume PSR applications, they are impractical for large-volume, high-speed PSR applications. A second option for generating airflow are syringe methods~\cite{Wu2020_Syringe}, which similarly suffer from limited flow rates in addition to having a small total capacity. 

Air reservoirs can achieve higher pressure gradients but inherently have a limited operable duration. For a fixed mass of air, choosing a reservoir involves a tradeoff between the size, weight, and reservoir pressure. Single-use CO$_2$ cartridges are compact options that contain pressures up to approximately 6000~kPa~\cite{Marchese2013, electronics_free}. However, this comes with practical challenges such as having to manually replace the canister after each use, sourcing components for the regulator that can withstand such a high inlet pressure, and preventing regulator ``freeze up". A lower-pressure or smaller reservoir can be made viable by incorporating a portable air compressor~\cite{Walsh} that refills the reservoir, but using a compressor increases system weight, size, and cost, and using a lower-pressure reservoir reduces dynamic performance. 

The phloSAR strikes a balance between these methods. It features an air reservoir at a moderate pressure (690~kPa), which facilitates dynamic performance while offering enough capacity to eliminate the need for an onboard compressor.

\subsection{Methods for Pressure Regulation}
Off-the-shelf pressure regulators make different trade-offs in performance, portability, and cost. An off-the-shelf regulator reduces complexity by handling feedback control internally, but requires a pressure source that satisfies the inlet specifications of the regulator. 

Alternatively, a pressure regulator can be custom-built by leveraging closed-loop control of flow valve arrays (binary solenoid valves and/or proportional valves) to achieve pressure regulation. This is a common strategy in existing untethered pneumatic systems~\cite{Shtarbanov2021_FlowIO, Marcin2019_Board, Holland2014_Toolkit, Oguntosin2018_LowCost ,Shrivastava2019_ProgrammableAir, Sinha2019_Board, Walsh}. Building a custom regulator with flow valves provides flexibility in that the designer selects regulator components jointly with the pressure source. Additionally, flow valves tend to be less bulky, heavy, and expensive  than off-the-shelf pressure regulators, and they can operate at higher inlet pressures, making them amenable to use with high-pressure air supplies.

PhloSAR implements pressure regulation via flow control but is intended for dynamic operation of large-volume PSRs, which have higher flow requirements than many of the systems in prior works.
\section{Design and Implementation}
We discuss design decisions for implementing a phloSAR for a specific PSR, but the design is highly customizable in that components can be readily swapped to achieve different performance specifications as needed. 

\subsection{Flow Modulation} 

Airflow in and out of the CV is modulated with proportional valves that have an affine relationship between flow and input current (Clippard ``EVP" EC-P-05-25-A0 and ``DVP" DV-PM-10-670100-V). The use of proportional valves, and the Clippard valves in particular, offers several benefits. First, they offer large maximum flow rates (EVP: 23.5 SLPM; DVP: 67.0 SLPM), which enables dynamic performance. Second, they allow the use of high inlet pressures (up to 690 kPa for the chosen valves). This allows the use of high-pressure reservoirs that are directly connected to the valve inlet, eliminating the need for intermediate components that would add weight and design complexity. Third, they are small and lightweight (EVP: 52~mm, 77~g; DVP: 64~mm, 139~g), which is critical for portability.



\subsection{Pressure Reservoir}
A high-pressure air reservoir allows for higher flow rates (Eq.~\ref{eq:ohm_analogy}) and a longer operation time due to the increased amount of air stored (ideal gas law). Larger reservoirs similarly allow longer operations times but there exists a trade off between portability and operation time. In the implemented prototype, we use a 2-liter bottle made from polyethylene terephthalate (PET). This reservoir choice is lightweight (70~g), low-cost ($<$\$1), and readily available (e.g. 2L soda bottle), and can withstand pressures over 900 kPa. We integrate a pneumatic fitting with the reservoir cap for routing tubing to the flow valves and to a manual on/off valve for refilling. The reservoir is easily refilled by opening the on/off valve and supplying compressed air. While it is the largest component of the implemented prototype (30~cm long), the large volume allows for more storage capacity. 

\subsection{Vacuum Source}
Passive venting to atmosphere or the use of vacuum pumps are simple solutions for decreasing CV pressure, but they suffer from limited pressure gradients and flow rates that reduce dynamic performance. Instead, the phloSAR design generates a vacuum with a Venturi pump (Coval VR09f14) that enables higher flow rates out of the CV. A Venturi pump is also advantageous in that it is pneumatically driven by the same pressure reservoir used for inflation. Specifically, high-velocity flow through the Venturi pump, from the inlet to the exhaust port, generates vacuum at a third port. This Venturi effect is used to actively draw air from CV, which is also released through the exhaust port.

\subsection{Operation and Controls}
The phloSAR controller coordinates commands to multiple valves to modulate airflow for CV pressure regulation. The first valve, the EVP, modulates flow from the reservoir into the CV. The second valve, the DVP, modulates flow from the reservoir into the Venturi pump. The third valve, a binary solenoid valve (Shenzhen Huamei Technology Co.), enables flow out of the CV and into the vacuum port of the Venturi pump. During inflation, the phloSAR modulates flow into the CV by actuating the EVP, and during deflation, the phloSAR modulates flow out of the CV by coordinating actuation of the DVP and solenoid valve. Actuating the solenoid valve alone allows passive venting from the CV, and actuating the DVP in tandem enables active deflation.

The valves are actuated with an L298N motor driver with commands from a Teensy LC microcontroller. The microcontroller uses pressure feedback from the CV and pressure reservoir (Honeywell pressure sensors ABPDANN030PGAA5 and ABPDANN015BGAA5) to determine the appropriate command. For low tracking errors, we use a PID controller. For high tracking errors, an on-off controller sends commands to maximize flow rate. The tracking error cutoff used to select between the two controllers (1~kPa) was determined empirically. During deflation, the phloSAR selects between passive venting and active deflation, only leveraging active deflation when additional speed is required. This improves air-use efficiency. 


\section{System Dynamics}
\label{sec:system-dynamics}
While we implement one example phloSAR, the design is highly customizable and intended to be adapted for other mobile PSRs. Here we provide models of system dynamics that give insight into phloSAR performance that can be used for designing other implementations of phloSAR. 

\subsection{Pressure Reservoir Discharge}
\label{sec:discharge-model}
Under constant operation, a pressure reservoir has a finite operation time before depletion. We provide a simple model based on the ideal gas law and Ohm's fluid law in Eq.~\ref{eq:ohm_analogy} to capture reservoir pressure over time:
\begin{equation}
    P_r(t) = P_{r,0} e^{-\frac{t}{\tau_{\mathrm{dis}}}},
    \label{eq:discharge}
\end{equation}
where $\tau_{\mathrm{dis}} = \frac{R_vV_r}{\alpha}$ and $\alpha = \frac{\rho R_uT}{M}$. $P_{r,0}$ is the initial pressure, $V_r$ is the reservoir volume, and $R_v$ is the flow resistance across the reservoir outlet valve. The model assumes the reservoir is discharging to atmosphere through a valve with constant flow resistance. In practice, the reservoir discharges to a CV, which is typically at a higher pressure than atmosphere, and the flow resistance can be modulated, but this model can be used to provide a conservative estimate of operation time by setting the model resistance to the minimum value for the implemented phloSAR. This model is also conservative because during real operation reservoir discharge halts when CV pressure stabilizes around the commanded pressure. In addition, reservoir discharge during CV deflation is more efficient than CV inflation because the phloSAR leverages both passive venting and active deflation, so the air discharged as motive fluid for the Venturi pump is, on average, less than the air vented from the CV.
\subsection{Inflation Speed}
\label{sec:fillup-model}
The time-rate at which the CV's pressure can be changed serves as a benchmark for the dynamic performance of a particular phloSAR setup. 
Assuming a high-pressure reservoir relative to the CV ($P_{r} \gg P_{cv}$), we derive the following from Eq.~\ref{eq:pressure-rate-to-flow} and \ref{eq:ohm_analogy}:\vspace{-2mm}
\begin{equation}
    \dot P_{cv} = \frac{\alpha P_r}{R_v V_{cv}}\\[-1mm]
    \label{eq:fillup-rate}
\end{equation}

This model is intended to aid in component selection for a phloSAR implementation. The size of the control volume $V_{cv}$ depends on the PSR application, but the reservoir pressure ($P_r$) and flow resistance ($R_v$) can be adjusted based on the choice of reservoir and proportional valve by the designer. $P_r$ decreases as the reservoir is depleted; we provide addition discussion in Sec.~\ref{sec:high-performance}.\\[-2mm]
\subsection{Frequency Response}
\label{sec:freq-response}
To model phloSAR frequency response, we consider a sinusoidal command and compute the maximum derivative:\vspace{-2mm}
\begin{equation}
    P_{\text{cmd}}(t) = A\sin(2\pi \omega t)+c \quad \rightarrow \quad \dot P_{\text{cmd}, \text{max}} = 2A\pi\omega,\\[-1mm]
    \label{eq:sine-command}
\end{equation}
where $A$ is the command amplitude and $c$ is a constant offset. For a phloSAR with a maximum inflation speed, $\dot P_{cv, \text{max}}$, larger than $\dot P_{\text{req}, \text{max}}$, the phloSAR tracks the command with unity gain. If $\dot P_{cv, \text{max}}$ is less than $\dot P_{\text{req}, \text{max}}$, we make a simplifying assumption that the phloSAR exhibits sinusoidal behavior with a maximum slope $\dot P_{cv, \text{max}}$. This can be used to solve for the output gain. This yields the following model for the phloSAR frequency response and cutoff frequency $\omega_c$:\vspace{-2mm}
\begin{equation}
    G(\omega) = \begin{cases} 
              1 & \omega\leq \omega_c \\[-1mm]
              \frac{\omega_c}{\omega} & \omega > \omega_c\\[-1mm]
    \end{cases}
\label{eq:frequency-response}
\end{equation}
\vspace{-2mm}
\begin{equation}
    \omega_c = \frac{\dot P_\text{cv,max}}{2\pi A} \quad \text{(Hz)}\\[-1mm]
    \label{eq:omega}
\end{equation}

This model shows that phloSAR bandwidth is affected by its PSR application. Specifically, $A$ is influenced by the magnitudes of pressure required by the PSR, and $\dot P_{cv}$ depends on the CV and phloSAR setup used (Eq.~\ref{eq:fillup-rate}).
\section{Experimental Verification}
Here we verify each model presented in Sec.~\ref{sec:system-dynamics}.
\subsection{Pressure Reservoir Discharge Verification}
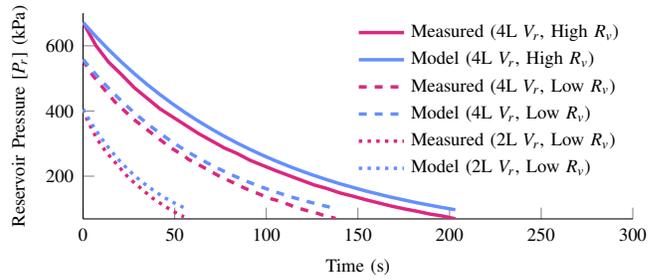
\begin{figure}[t]
    \vspace{.5mm}
%
%
\definecolor{mycolor1}{rgb}{0.86275,0.14902,0.49804}%
\definecolor{mycolor2}{rgb}{0.39216,0.56078,1.00000}%
\begin{tikzpicture}[font=\scriptsize]

\begin{axis}[%
width=3.5in,
height=1.7in,
xmin=0,
xmax=300,
xlabel={Time (s)},
ymin=68.94757,
ymax=700,
ylabel={Reservoir Pressure [$P_r$] (kPa)},
axis background/.style={fill=white},
axis x line*=bottom,
axis y line*=left,
legend style={legend cell align=left, align=left, at={(1.0,1.0)}, anchor=north east, fill=none, draw=none}
]
\addplot [color=mycolor1, line width=1.3pt]
  table[row sep=crcr]{%
0	671.68722694\\
7.005	600.80912498\\
14.01	549.5121329\\
21.015	511.5909694\\
28.02	471.25664095\\
42.03	406.51487272\\
63.045	330.74149329\\
77.0550000000001	286.75294363\\
84.061	269.65394627\\
91.066	248.9007277\\
105.076	217.11589793\\
126.091	173.12734827\\
133.096	162.16468464\\
140.101	148.71990849\\
147.106	138.92935355\\
154.111	127.96668992\\
168.121	108.38558004\\
175.126	99.83608136\\
182.132	92.52763894\\
189.137	83.97814026\\
196.142	77.84180653\\
203.147	69.29230785\\
};
\addlegendentry{Measured (4L $V_r$, High $R_v$)}

\addplot [color=mycolor2, line width=1.3pt]
  table[row sep=crcr]{%
0	671.68722694\\
7.005	628.345680866648\\
14.01	587.800807918356\\
21.015	549.872148898876\\
28.02	514.390888990859\\
35.025	481.199106386214\\
42.03	450.149069399643\\
49.035	421.102578936897\\
56.04	393.930353391166\\
63.045	368.511453229886\\
70.05	344.732742710878\\
77.0550000000001	322.488386331996\\
84.061	301.67650516452\\
91.066	282.210411983916\\
98.071	264.000395352957\\
105.076	246.965405197344\\
112.081	231.029621310775\\
119.086	216.122115890482\\
126.091	202.176537848138\\
133.096	189.130817491037\\
140.101	176.926890259128\\
154.111	154.830648649639\\
168.121	135.493986957877\\
182.132	118.571134655871\\
196.142	103.76289134458\\
203.147	97.0674474600752\\
};
\addlegendentry{Model (4L $V_r$, High $R_v$)}

\addplot [color=mycolor1, dashed, line width=1.3pt]
  table[row sep=crcr]{%
0	558.06163158\\
2.505	529.93102302\\
7.51499999999999	493.25091578\\
17.535	427.26809129\\
27.5549999999999	374.73004295\\
47.595	287.99399989\\
55.111	262.34550385\\
57.616	251.31389265\\
62.626	236.69700781\\
75.151	197.60373562\\
77.6560000000001	191.46740189\\
87.676	164.57784959\\
92.687	152.3741297\\
95.192	147.47885223\\
97.697	141.3425185\\
100.202	136.5161886\\
102.707	130.37985487\\
105.212	126.72563366\\
112.727	112.03980125\\
115.232	105.97241509\\
117.737	102.24924631\\
120.242	97.35396884\\
127.758	86.39130521\\
132.768	76.60075027\\
135.273	74.18758532\\
137.778	70.53336411\\
};
\addlegendentry{Measured (4L $V_r$, Low $R_v$)}

\addplot [color=mycolor2, dashed, line width=1.3pt]
  table[row sep=crcr]{%
0	558.06163158\\
7.51499999999999	508.389134793386\\
15.03	463.13793629605\\
22.545	421.914500835541\\
30.0599999999999	384.360321330952\\
37.575	350.148801051088\\
45.09	318.982413306774\\
52.606	290.586509259698\\
60.121	264.721700373681\\
67.636	241.159091752963\\
75.151	219.693766899423\\
82.6660000000001	200.139048723485\\
90.182	182.322614379315\\
97.697	166.094264382841\\
105.212	151.310383271945\\
112.727	137.842400344006\\
120.242	125.573188843546\\
130.263	110.894582468587\\
137.778	101.023968759258\\
};
\addlegendentry{Model (4L $V_r$, Low $R_v$)}

\addplot [color=mycolor1, dotted, line width=1.3pt]
  table[row sep=crcr]{%
0	405.27381646\\
2.505	369.83476548\\
7.51499999999999	314.88355219\\
12.525	274.54922374\\
25.051	193.88056684\\
37.576	134.03407608\\
45.091	105.97241509\\
52.606	81.49602774\\
55.111	75.35969401\\
};
\addlegendentry{Measured (2L $V_r$, Low $R_v$)}

\addplot [color=mycolor2, dotted, line width=1.3pt]
  table[row sep=crcr]{%
0	405.27381646\\
5.00999999999999	357.904649750338\\
10.02	316.072080431466\\
15.03	279.128980576148\\
20.04	246.503859787683\\
25.051	217.686623748174\\
30.061	192.243001308261\\
35.071	169.773277363893\\
40.081	149.92985705971\\
45.091	132.405773081494\\
50.101	116.929937032663\\
55.111	103.262945838829\\
};
\addlegendentry{Model (2L $V_r$, Low $R_v$)}

\end{axis}
\end{tikzpicture}%
    \vspace{-9mm}
    \caption{Characterization of pressure reservoir discharge (singular trials) through a valve with constant flow resistance. We compare experimental data to the model presented in Sec.~\ref{sec:discharge-model} for different initial reservoir pressures, reservoir sizes, and flow resistances. The tests shown represent the conditions with shortest and longest duration and one intermediate example.}
    \label{fig:discharge}
    \vspace{-5.5mm}
\end{figure}

Equation~\ref{eq:discharge} is verified with discharge tests that use the following procedure: fill a reservoir to an initial pressure, open an outlet valve at constant flow resistance, and monitor the reservoir pressure ($P_r$) over time. We performed discharge tests with varied initial reservoir pressure ($P_{r,0}$), reservoir volumes ($V_{cv}$), and flow resistances ($R_v$), and compared the measured and modeled pressures over time with  
the nRMSE shown in Fig.~\ref{fig:discharge}. The nRMSE for the displayed tests are in the range 4.1-5.7\% (normalized with respect to $P_{r, 0}$), which suggests that reservoir discharge is captured well by Eq.~\ref{eq:discharge}. 

\begin{figure}[b]
    \vspace{-8mm}
    \centering
    \input{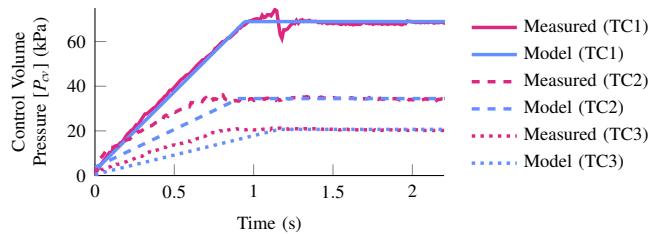}
    \vspace{-8mm}
    \caption{Comparing PhloSAR step response with the model (piecewise linear with an initial inflation speed given by Eq.~\ref{eq:fillup-rate}). The pressure command, phloSAR reservoir pressure, and control volume size were varied (TC1, TC2, TC3) to yield different inflation speeds.}
    \label{fig:fillup}
\end{figure}
\subsection{Inflation Speed Verification}

We verify Eq~\ref{eq:fillup-rate} by analyzing the step response of the phloSAR prototype. We model the response as a piece-wise linear function, in which the pressure increases linearly according to Eq.~\ref{eq:fillup-rate} and then remains constant after reaching the target pressure. We collect experimental data for 3 different test conditions (TC) to show varying inflation speeds: 
\begin{itemize}[leftmargin=2.0cm,labelsep=0.5cm]
    \item[TC1:] $\Delta P_{cv}$ = 69 kPa, $V_{cv}$ = 0.5 L, $P_r$ = 689 kPa
    \item[TC2:] $\Delta P_{cv}$ = 34 kPa, $V_{cv}$ = 1 L, \hspace{2.5mm}$P_r$ = 689 kPa 
    \item[TC3:] $\Delta P_{cv}$ = 21 kPa, $V_{cv}$ = 1 L, \hspace{2.5mm}$P_r$ = 345 kPa
\end{itemize}
The CVs used were rigid containers.

From Fig.~\ref{fig:fillup}, we see that a piecewise linear function for the phloSAR step response captures the behavior well. The nRMSE between the measured and modeled time series, normalized by the commanded pressure, is 2.3\%, 10.9\%, and 12.9\%, for TC1, TC2, and TC3 respectively. For each test, we also compute the average rate of pressure increase and compare it to the rate of change given by Eq.~\ref{eq:fillup-rate}. The normalized errors for the three tests, normalized by the model speed, are 0.24\%, 44.8\%, and 45.0\%, for TC1, TC2, and TC3 respectively. In future work, we will explore what causes deviation from the model for larger control volumes. We also note the prioritization of inflation speed, which sometimes caused overshoot (e.g. TC1). We anticipate this could be resolved with further controller design and tuning.

\subsection{Frequency Response Verification}
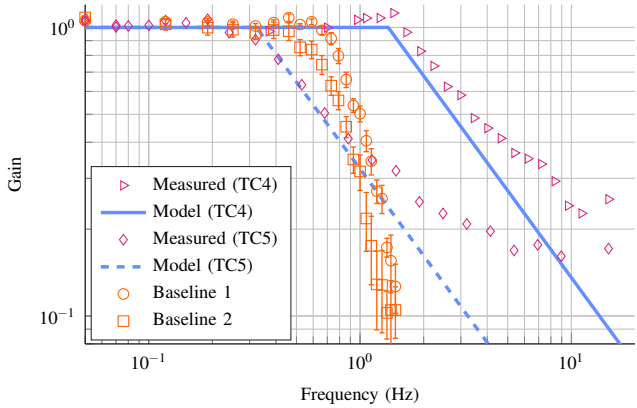
\begin{figure}[t]
    \vspace{1.5mm}
%
%
\definecolor{mycolor1}{rgb}{0.86275,0.14902,0.49804}%
\definecolor{mycolor2}{rgb}{0.39216,0.56078,1.00000}%
\definecolor{mycolor3}{rgb}{0.99608,0.38039,0.00000}%
\begin{tikzpicture}[font=\scriptsize]

\begin{axis}[%
width=3.5in,
height=2.4in,
xmode=log,
xmin=0.05,
xmax=20,
xminorticks=true,
xlabel={Frequency (Hz)},
ymode=log,
ymin=0.08,
ymax=1.2,
yminorticks=true,
ylabel={Gain},
axis background/.style={fill=white},
axis x line*=bottom,
axis y line*=left,
xmajorgrids,
xminorgrids,
ymajorgrids,
yminorgrids,
legend style={at={(0.01,0.01)}, anchor=south west, legend cell align=left, align=left, draw=white!15!black}
]
\addplot [color=mycolor1, only marks, mark=triangle, mark options={solid, rotate=270, mycolor1}]
  table[row sep=crcr]{%
0.05	1.05614939772437\\
0.37	0.969148635039553\\
0.7	0.998434391884949\\
0.97	1.06302358827315\\
1.07	1.07752626269168\\
1.24	1.07817905486979\\
1.44	1.12141985246722\\
1.66	0.961565456326516\\
1.93	0.82703667878588\\
2.23	0.735368080687163\\
2.58	0.623556740543292\\
2.99	0.582296709106754\\
3.46	0.485065133598085\\
4.01	0.447681053843684\\
4.64	0.413012154269148\\
5.38	0.36719652806081\\
6.23	0.350927990286323\\
7.21	0.336051457127794\\
8.35	0.29307894867314\\
9.66	0.241103011897592\\
11.19	0.226745627151045\\
15	0.253800614814173\\
};
\addlegendentry{Measured (TC4)}

\addplot [color=mycolor2, line width=1.3pt]
  table[row sep=crcr]{%
0.05	1\\
1.3587 1\\
19.275	0.070475971447446\\
};
\addlegendentry{Model (TC4)}

\addplot [color=mycolor1, only marks, mark=diamond, mark options={solid, mycolor1}]
  table[row sep=crcr]{%
0.05	1.06441177129442\\
0.07	1.00297311148727\\
0.08	1.01231905810974\\
0.05	1.04708721779335\\
0.07	1.01898715850179\\
0.1	1.017187191168\\
0.12	1.05619393878588\\
0.15	1.03710574472309\\
0.19	1.06736304803288\\
0.24	0.959409266676892\\
0.32	0.90696031844578\\
0.41	0.775920397899175\\
0.53	0.632768709382185\\
0.68	0.505061202238949\\
0.88	0.411387959226437\\
1.14	0.34706670953118\\
1.48	0.31854330114423\\
1.91	0.248631928137051\\
2.47	0.226528789360531\\
3.2	0.208397332019296\\
4.14	0.196801961780685\\
5.36	0.168645784044991\\
6.93	0.176354450958977\\
8.96	0.161089281605292\\
15	0.171128204791885\\
};
\addlegendentry{Measured (TC5)}

\addplot [color=mycolor2, dashed, line width=1.3pt]
  table[row sep=crcr]{%
0.05	1\\
0.3234	1\\
4.8818	0.0662707389886347\\
};
\addlegendentry{Model (TC5)}

\addplot [color=mycolor3, only marks, mark=o, mark options={solid, mycolor3}]
 plot [
 error bars/.cd, y dir=both, y explicit, error bar style={line width=0.5pt}, error mark options={line width=0.5pt, mark size=1.0pt, rotate=90}
 ]
 table[row sep=crcr, y error plus index=2, y error minus index=3]{%
0.05	1.05517299227879	0.0395691151111827	0.0395691151111827\\
0.12	1.04688369886993	0.0176044251278276	0.0176044251278276\\
0.19	1.04489111598956	0.0295190276271161	0.0295190276271161\\
0.25	1.02653019896169	0.0255946904530766	0.0255946904530766\\
0.32	1.01136010422229	0.0201129438258528	0.0201129438258528\\
0.39	1.03920064756387	0.0232509261311153	0.0232509261311153\\
0.46	1.08118421951893	0.0315422388702371	0.0315422388702371\\
0.52	1.02322600505447	0.00697725821404645	0.00697725821404645\\
0.59	1.04636146888747	0.0238325126055803	0.0238325126055803\\
0.66	0.977096955027861	0.031538254612486	0.031538254612486\\
0.73	0.915451328334196	0.0433627775388444	0.0433627775388444\\
0.79	0.797350558030404	0.0479984389163795	0.0479984389163795\\
0.86	0.659516091629339	0.0402185948819429	0.0402185948819429\\
0.93	0.536618561660502	0.0293361458305409	0.0293361458305409\\
1	0.502350889272632	0.0314991160691647	0.0314991160691647\\
1.07	0.40475630531403	0.0337765039525681	0.0337765039525681\\
1.13	0.344056526953189	0.03591138399607	0.03591138399607\\
1.2	0.270795225603153	0.0254055744684178	0.0254055744684178\\
1.27	0.255029706349374	0.0285692866919804	0.0285692866919804\\
1.34	0.172904342345094	0.0129608609455257	0.0129608609455257\\
1.4	0.155404329732854	0.0352508656973886	0.0352508656973886\\
1.47	0.126576823787718	0.0245575147785439	0.0245575147785439\\
};
\addlegendentry{Baseline 1}

\addplot [color=mycolor3, only marks, mark=square, mark options={solid, mycolor3}]
 plot [
 error bars/.cd, y dir=both, y explicit, error bar style={line width=0.5pt}, error mark options={line width=0.5pt, mark size=1.0pt, rotate=90}]
 table[row sep=crcr, y error plus index=2, y error minus index=3]{%
0.05	1.08261492453681	0.0264555063094568	0.0264555063094568\\
0.12	1.02489112176478	0.0454459163768522	0.0454459163768522\\
0.19	0.998306931532193	0.058124361547918	0.058124361547918\\
0.25	0.982324876837008	0.0568041248800605	0.0568041248800605\\
0.32	0.958345889412245	0.0492359267251854	0.0492359267251854\\
0.39	0.97938702569953	0.0639284646066212	0.0639284646066212\\
0.46	0.969538133782409	0.0684793492631252	0.0684793492631252\\
0.52	0.85311260532476	0.0551183577759792	0.0551183577759792\\
0.59	0.838125646658215	0.0584696936007677	0.0584696936007677\\
0.66	0.743293388975451	0.0553992966591987	0.0553992966591987\\
0.73	0.628310112284367	0.0464530261540583	0.0464530261540583\\
0.79	0.558488573867338	0.039908953817063	0.039908953817063\\
0.86	0.452736053516813	0.0377996496240016	0.0377996496240016\\
0.93	0.348640855052496	0.036362082836862	0.036362082836862\\
1	0.31678990496523	0.0444879721528861	0.0444879721528861\\
1.07	0.217685110727333	0.0501986273012588	0.0501986273012588\\
1.13	0.174967279435515	0.0466306210893015	0.0466306210893015\\
1.2	0.128587736446803	0.0390898718148217	0.0390898718148217\\
1.27	0.128008021790816	0.0405814371965089	0.0405814371965089\\
1.34	0.102468556821197	0.0193552370212134	0.0193552370212134\\
1.4	0.104786790824304	0.0277371724233249	0.0277371724233249\\
1.47	0.104897858175145	0.021255839908608	0.021255839908608\\
};
\addlegendentry{Baseline 2}

\end{axis}

\end{tikzpicture}%
    \vspace{-4mm}
    \caption{Comparison of phloSAR closed-loop frequency response to the model presented in Sec.~\ref{sec:freq-response}. Test condition TC4 is a phloSAR and control volume within expected operating conditions, and test condition TC5 is a more difficult condition (lower reservoir pressure, larger control volume, and higher-amplitude pressure command). The frequency response of a baseline (off-the-shelf pressure regulator) is overlaid.  Baselines 1 and 2 correspond to TC4 and TC5, except the baseline regulator is supplied with its maximum recommended inlet pressure in both cases. Error bars represent 1 standard deviation, and bars not shown are within the bounds of the marker.}
    \vspace{-6mm}
    \label{fig:freq_response}
\end{figure}
We verify Eq.~\ref{eq:frequency-response} by analyzing the phloSAR's pressure output in response to sinusoidal pressure commands with offset $c$ such that $P_{\text{cmd, min}} = 0$ kPa (Fig.~\ref{fig:freq_response}). We consider two test conditions, one with expected operating conditions (TC4) and a second with more extreme conditions (TC5), specifically, a higher amplitude pressure command, larger CV, and lower pressure reservoir:
\begin{itemize}[leftmargin=2.0cm,labelsep=0.5cm]
    \item[TC4:] $A$ = 21 kPa, $V_{cv}$ = 0.5 L, $P_r$ = 689 kPa
    \item[TC5:] $A$ = 34 kPa, $V_{cv}$ = 1 L, \hspace{1.2mm} $P_r$ = 345 kPa 
\end{itemize}

For each condition, we measure the phloSAR response for a wide range of command frequencies. To compute the gain for a given trial, a Fast Fourier Transform (FFT) is applied to the phloSAR's pressure time response, and the magnitude at the command frequency is isolated. We used a fixed pressure source of 690~kPa to eliminate the effect of changing $P_r$ on frequency response.

We compare the measured frequency response to the model from Eq.~\ref{eq:frequency-response}, using a measured value for $\dot P_\text{cv,max}$. Both test conditions show qualitative agreement between the modeled and measured frequency response. Although the roll-off rate deviates from the model for the highest frequencies ranges, this is beyond the intended operating bandwidth as these frequencies are much higher than the cutoff frequency. The modeled and measured cutoff frequency for TC4 is 1.35 and 1.6 Hz, respectively, and the modeled and measured cutoff frequency for TC5 is 0.32 and 0.22 Hz, respectively.

Figure~\ref{fig:freq_response} also provides a qualitative ``baseline" comparison with one of the off-the-shelf regulators from Fig.~\ref{fig:compass} (Proportion Air QB3TANKKZP6PSG) commonly used to actuate vine robots~\cite{CoadRAM2020, multisegment_vine}. However, the baseline regulator is not suited for untethered operation because it does not have an integrated pressure source. Baseline 1 and 2 correspond to TC4 and TC5, except the parameter for reservoir pressure is not applicable. Instead, the baseline regulator is supplied with its maximum recommended inlet pressure. The phloSAR has better performance than the baseline at expected operating conditions (TC4) and worse performance at more extreme conditions (TC5); this is expected when operating outside intended conditions.

\begin{figure}[t]
    \vspace{1mm}
    \input{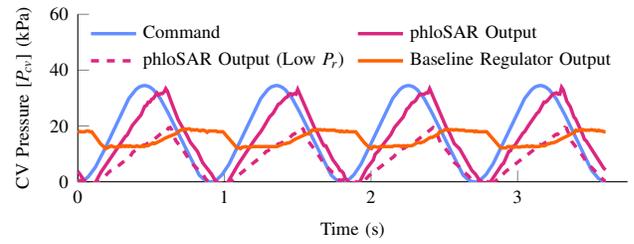}
    \vspace{-3.5mm}
    \caption{Time series data for a 1.2~Hz pressure command. Comparison of phloSAR output with a ``baseline" off-the-shelf regulator, as well as phloSAR output when the reservoir pressure is too low. The peak-to-peak measure of the command is 34~kPa, and the approximate peak-to-peak measure of the phloSAR, low $P_r$ phloSAR, and baseline is 32, 20, and 6~kPa, respectively, demonstrating the higher dynamic performance of the phloSAR compared to the baseline.}
    \vspace{-4mm}
    \label{fig:freq_time}
\end{figure}

Figure~\ref{fig:freq_time} provides time-cropped segments from trials with a 1.2~Hz pressure command to visualize the difference in performance for a phloSAR and the baseline regulator. We also overlay the response of a phloSAR with low reservoir pressure to show behavior when operating outside of intended conditions. The peak-to-peak measure of the command is 34~kPa, and the approximate peak-to-peak measure of the phloSAR, low $P_r$ phloSAR, and baseline is 32, 20, and 6~kPa, respectively. This demonstrates the higher dynamic performance of the phloSAR compared to the baseline. As a result, phloSAR exhibits the best tracking performance, although when the reservoir pressure is too low (Low $P_r$), the inflation speed is not high enough to track the command. The baseline suffers from slow response times and low inflation speeds leading to poor tracking performance.
\section{Influence of Design Parameters on System~Operation}
Here we synthesize findings from our work to serve as a guideline when implementing a phloSAR.
\subsection{Designing for Performance}
\label{sec:regulation-performance}
The high-level insights from Secs.~\ref{sec:fillup-model} and \ref{sec:freq-response} are that maximizing phloSAR dynamic ability can be achieved through:
\begin{enumerate}
    \item Maximizing starting reservoir pressure ($P_{r, 0}$)
    \item Minimizing minimum flow resistance ($R_{v, \text{min}}$) when sourcing proportional flow valves
    \item Minimizing volume ($V_{cv}$) and pressure difference between ``on" and ``off" states ($\Delta P_{cv}$) of the CV (e.g. PSR actuators).
\end{enumerate}
However, this needs to be considered jointly with cost, size, and weight considerations of the phloSAR, and performance requirements of the PSR.
\subsection{Designing for Operation Duration}
\label{sec:high-performance}
The discharge model in Sec.~\ref{sec:discharge-model} gives a sense of operable duration, and we supplement this with a metric for the number of CV inflation-deflation cycles ($n_{\text{cycles}}$) that can be achieved for a given phloSAR and setup. 

Given a target inflation speed, $\dot P_{cv, d}$, and the change in pressure during inflation, $\Delta P_{cv}$, there is a minimum reservoir pressure for the phloSAR to satisfy the target inflation speed (Eq.~\ref{eq:fillup-rate}). For a specific reservoir volume and initial pressure, we can compute the quantity of air that can be supplied before reaching this minimum pressure. Additionally, for a specific CV ($V_{cv}$), we can compute the quantity of air required per inflation-deflation cycle. These quantities can be related to yield
\begin{equation}
    n_\text{cycles} = \frac{1}{2}\left(   \frac{P_{r, 0}}{V_{cv}} - \frac{R_{v\text{,min}} \dot P_{cv,d}} {\alpha } \right) \frac{V_{r}}{\Delta P_{cv}},
    \label{eq:high-performance-fillups}
\end{equation} 
which gives the number of inflation-deflation cycles possible while satisfying the PSR's dynamic specification. For simplicity, the model conservatively assumes the quantity of air required by the Venturi pump during deflation is the same as the air required by the CV during inflation.
\section{Demonstration}
\label{sec:demo}
The aim of this work is to enable untethered operation of soft robots. To demonstrate this, we use the phloSAR prototype to inflate the bending actuator of a vine robot mounted to an aerial vehicle. We envision that these ``Flying Vines" could be used in hard-to-access areas at high altitude, for example in bridge inspection~\cite{bridge_inspection} or canopy sampling~\cite{canopy_sampling}.

Fig.~\ref{fig:splash} shows the proof-of-concept assembly in flight and a visualization of mass distribution. The aerial vehicle features a Lumenier QAV-PRO Frame with 12.7~cm propellers and RaceSpec 2300~kv brushless motors (RaceSpec RS2205). The vehicle consumes approximately 1-2~kW of power with a payload of approximately 1.5~kg; however, larger aerial vehicles with longer flight duration may also be used. Meanwhile, the phloSAR consumes 3~W and weighs 0.84~kg, a comparable mass to that of an off-the-shelf regulator. For example, the QB3 regulator commonly used for vine robots is 0.68~kg, but it does not include a pressure source or electronic hardware for sending pressure commands. During testing, the phloSAR inflated a 0.1 L bending actuator to 20.7~kPa at 0.55~Hz. We did not test the maximum number of inflation cycles, but about 30 cycles were performed during testing, and Eq.~\ref{eq:high-performance-fillups} predicts more than 300 cycles. 
\section{Conclusion}
We present a portable air supply and pressure regulator that enables untethered operation of soft robots with high-flow requirements. We provide models (verified on hardware) that aid in selecting system components based soft robot requirements. We implement a phloSAR for a ``vine" robot, and demonstrate that it is portable enough to be mounted on an aerial vehicle. Our findings show that phloSARs can enable untethered operation of pneumatic soft robots, which can lead to novel soft robot applications. In the future, we plan to develop higher-fidelity dynamic models for more accurate component specification and explore options for onboard reservoir re-pressurization.





\section*{Acknowledgement}
The authors thank Ian Scholl for helpful discussions, and Jun En Low and Keiko Nagami for assistance with drones.

\bibliographystyle{IEEEtran}
\bibliography{CHARMBib, paper}

\end{document}